# Remote Sensing Image Scene Classification: Benchmark and State of the Art

*This paper reviews the recent progress of remote sensing image scene classification, proposes a large-scale benchmark dataset, and evaluates a number of state-of-the-art methods using the proposed dataset.*

Gong Cheng, Junwei Han, *Senior Member, IEEE*, and Xiaoqiang Lu, *Senior Member, IEEE*

***Abstract*—Remote sensing image scene classification plays an important role in a wide range of applications and hence has been receiving remarkable attention. During the past years, significant efforts have been made to develop various datasets or present a variety of approaches for scene classification from remote sensing images. However, a systematic review of the literature concerning datasets and methods for scene classification is still lacking. In addition, almost all existing datasets have a number of limitations, including the small scale of scene classes and the image numbers, the lack of image variations and diversity, and the saturation of accuracy. These limitations severely limit the development of new approaches especially deep learning-based methods. This paper first provides a comprehensive review of the recent progress. Then, we propose a large-scale dataset, termed "NWPU-RESISC45", which is a publicly available benchmark for REmote Sensing Image Scene Classification (RESISC), created by Northwestern Polytechnical University (NWPU). This dataset contains 31,500 images, covering 45 scene classes with 700 images in each class. The proposed NWPU-RESISC45 (i) is large-scale on the scene classes and the total image number, (ii) holds big variations in translation, spatial resolution, viewpoint, object pose, illumination, background, and occlusion, and (iii) has high within-class diversity and between-class similarity. The creation of this dataset will enable the community to develop and evaluate various data-driven algorithms. Finally, several representative methods are evaluated using the proposed dataset and the results are reported as a useful baseline for future research.**

This work was supported in part by the National Science Foundation of China under Grants 61401357, 61522207, 61473231, and the Fundamental Research Funds for the Central Universities under Grant 3102016ZY023 (*Corresponding author: Junwei Han*).

Gong Cheng and Junwei Han are with the School of Automation, Northwestern Polytechnical University, Xi'an 710072, Shaanxi, P.R. China. Email: junweihan2010@gmail.com.

Xiaoqiang Lu is with the Center for OPTical IMagery Analysis and Learning (OPTIMAL), State Key Laboratory of Transient Optics and Photonics, Xi'an Institute of Optics and Precision Mechanics, Chinese Academy of Sciences, Xi'an 710119, Shaanxi, P.R. China.

## I. Introduction

The currently available instruments (e.g., multi/hyperspectral [1], synthetic aperture radar [2], etc.) for earth observation [3, 4] generate more and more different types of airborne or satellite images with different resolutions (spatial resolution, spectral resolution, and temporal resolution). This raises an important demand for intelligent earth observation through remote sensing images, which allows the smart identification and classification of land use and land cover (LULC) scenes from airborne or space platforms [3]. Remote sensing image scene classification, being an active research topic in the field of aerial and satellite image analysis, is to categorize scene images into a discrete set of meaningful LULC classes according to the image contents. During the past decades, remarkable efforts have been made in developing various methods for the task of remote sensing image scene classification because of its important role for a wide range of applications, such as natural hazards detection [5-7], LULC determination [8-43], geospatial object detection [27, 44-52], geographic image retrieval [53-63], vegetation mapping [64-68], environment monitoring, and urban planning.

In the early 1970s, the spatial resolution of satellite images (such as Landsat series) is low and hence, the pixel sizes are typically coarser than, or at the best, similar in size to the objects of interest [69]. Most of the methods for image analysis using remote sensing images developed since the early 1970s are based on per-pixel analysis, or even sub-pixel analysis for this conversion [69-72]. With the advances of remote sensing technology, the spatial resolution is gradually finer than the typical object of interest and the objects are generally composed of many pixels, which has significantly increased the within class variability and single pixels do not come isolated but are knitted into an image full of spatial patterns. In this case, it is difficult or sometimes impoverished to categorize scene images at pixel level purely.

Having identified an increasing dissatisfaction with pixel level image classification paradigm, Blaschke and Strobl [70] in 2001 raised a critical question "What's wrong with pixels?" to conclude that researchers should focus on the spatial patterns created by pixels rather than the statistical analysis of single pixels. Afterwards, a new paradigm of object based image analysis (OBIA) [71] or geographic object based image analysis (GEOBIA) [73] was raised for the object level delineation and analysis of remote sensing images rather than individual pixels. Here, the term "objects" represents meaningful semantic entities or scene components that are distinguishable in an image (e.g., a house, tree or vehicle in a 1:3000 scale color airphoto). The core task of OBIA and GEOBIA is the production of a set of nonoverlapping segments (or polygons), that is, the partitioning of a scene image into meaningful geographically based objects or superpixels that share relatively homogeneous spectral, color, or texture information. Due to the superiority compared to pixel-level approaches, object-level methods have dominated the task of remote sensing image analysis for decades [70-80]. Although pixel- and object-level classification methods have demonstrated impressive performance for some typical land use identification tasks, but pixels, or even superpixels, carry little semantic meanings. For semantic-level understanding of the meanings and contents of remote sensing images, we take eight images from the very popular UC Merced land use dataset [38] as examples, as shown in Figure 1 (a)-(h) (dense residential *v.s.* medium residential, freeway *v.s.* runway, river *v.s.* golf course, intersection *v.s.* overpass), such pixel- and object-level methods are distinctly not enough to identify them correctly. Fortunately, with the rapid development of machine learning theories, recent years have witnessed the emergence of another research flow, that is, semantic-level remote sensing image scene classification which aims to label each scene image with a specific semantic class [9-11, 15, 16, 18, 23, 24, 26, 28, 33, 36-38, 53, 55, 59, 81-87]. Here, a scene image usually refers to a local image patch manually extracted from large scale remote sensing images that contain explicit semantic classes (e.g., commercial area, industrial area, and residential area).

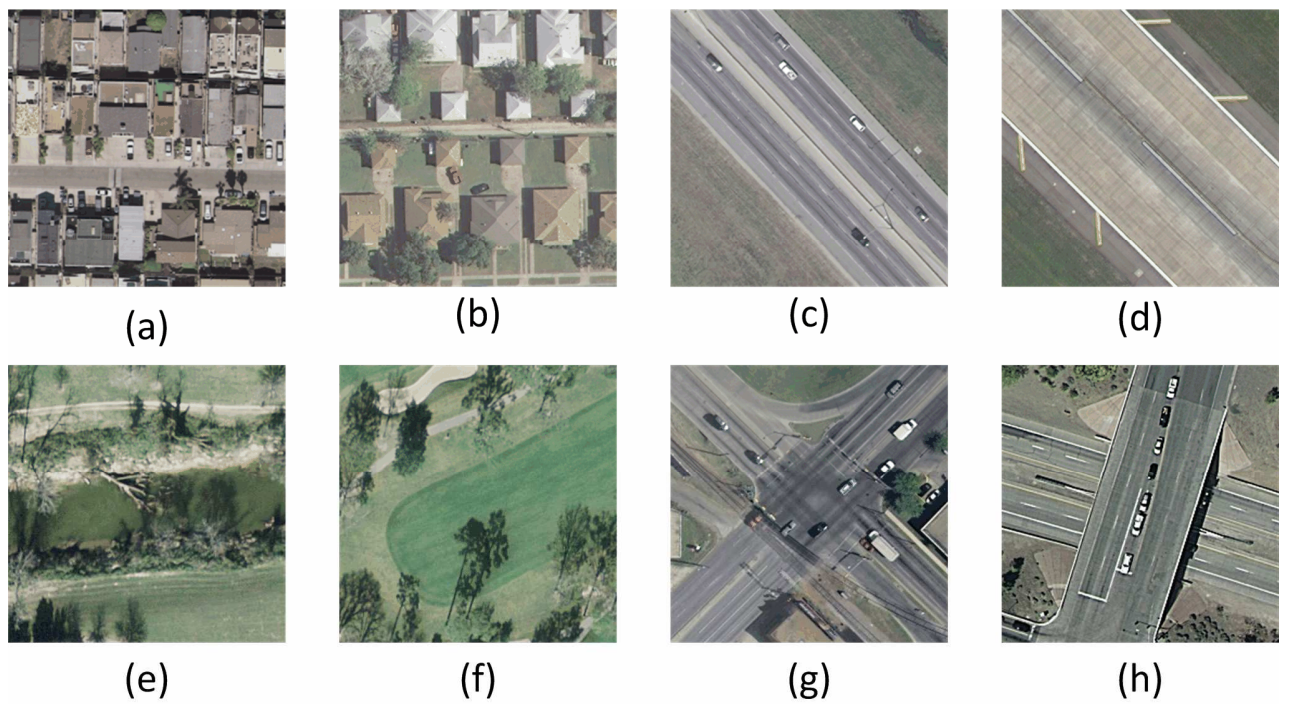

Figure 1. Eight scene images from the popular UC Merced land use dataset: (a) dense residential, (b) medium residential, (c) freeway, (d) runway, (e) river, (f) golf course, (g) intersection, and (h) overpass.

Since the emergence of UC Merced land use dataset [38], the first publicly available high resolution remote sensing image dataset for scene classification, significant efforts have been made in developing publicly available datasets [9, 11, 17, 33, 38, 82] (as summarized in Table 1) or presenting various methods [9-11, 15, 16, 18, 24, 26, 28, 33, 36-38, 53, 55, 59, 81-87] for the task of remote sensing image scene classification. However, a deep review of the literatures concerning datasets and methods for scene classification is still lacking. Besides, almost all existing datasets have a number of limitations, such as the small scale of scene classes and images per class, the lack of image variations and diversity, and the saturation of accuracy (e.g., almost 100% classification accuracy on the most popular UC Merced dataset [38] with deep ConvNets features [82]). These limitations severely limit the development of new data driven algorithms especially deep learning based methods.

With this in mind, this paper first provides a comprehensive review of the recent progress in this field. Then, we propose a large-scale benchmark dataset, named "NWPU-RESISC45"[1], which is a freely and publicly available benchmark used for REmote Sensing Image Scene Classification (RESISC), created by Northwestern Polytechnical University (NWPU). This new dataset has a total number of 31,500 images, covering 45 scene classes with 700 images in each class. We highlight three key features of the proposed NWPU-RESISC45 dataset when comparing it with other existing scene datasets [9, 11, 17, 33, 38, 82]. First, it is large-scale on the number of scene classes, the number of image per class, and the total number of images. Second, for each scene class, our dataset possesses big image variations in translation, spatial resolution, viewpoint, object pose, illumination, background, and occlusion, etc. Third, it has high within class diversity and between class similarity.

To sum up, the main contribution of this paper is threefold.

*1) This paper provides a comprehensive review of the recent progress in this field.* Covering about 170 publications we review existing publicly available benchmark datasets and three main categories of approaches for remote sensing image scene classification, including handcrafted feature based methods, unsupervised feature learning based methods, and deep feature learning based methods.

*2) We propose a large-scale, publicly available benchmark dataset by analyzing the limitations of existing datasets.* To the best of our knowledge, this dataset is of the largest scale on the number of scene classes and the total number of images. The creation of this dataset will enable the community to develop and evaluate various new data-driven algorithms, at a much larger scale, to further improve the state-of-the-arts.

*3) We investigate how well the current state-of-the-art scene classification methods perform on NWPU-RESISC45 dataset.* Current methods have only been tested on small datasets. It is unclear how they compare to each other on a larger dataset with a large number of scene classes. Thus we evaluate a number of representative methods including deep learning based methods for the task of scene classification using the proposed dataset and the results are reported as a useful performance baseline.

The rest of the paper is organized as follows. Section II reviews several publicly available datasets for remote sensing image scene classification. Section III surveys three categories of approaches in this domain. Section IV provides details on our dataset. A benchmarking of state-of-the-art scene classification methods using our dataset is given in Section V. Finally, conclusions are drawn in Section VI.

[1] http://www.escience.cn/people/JunweiHan/NWPU-RESISC45.html
https://1drv.ms/u/s!AmgKYzARBl5ca3HNaHIlzp_IXjs (OneDrive)
http://pan.baidu.com/s/1mifR6tU (BaiduWangpan)

## II. A Review on Remote Sensing Image Scene Classification Datasets

In the past years, several publicly available high resolution remote sensing image datasets [9, 11, 17, 33, 38, 82] have been introduced by different groups to perform research for scene classification and to evaluate different methods in this field. We will briefly review them in this section.

### A. UC Merced Land-Use Dataset

The UC Merced land-use dataset [38] is composed of 2,100 overhead scene images divided into 21 land-use scene classes. Each class consists of 100 aerial images measuring 256×256 pixels, with a spatial resolution of 0.3 m per pixel in the red green blue color space. This dataset was extracted from aerial orthoimagery downloaded from the United States Geological Survey (USGS) National Map of the following US regions: Birmingham, Boston, Buffalo, Columbus, Dallas, Harrisburg, Houston, Jacksonville, Las Vegas, Los Angeles, Miami, Napa, New York, Reno, San Diego, Santa Barbara, Seattle, Tampa, Tucson, and Ventura. The 21 land-use classes are: agricultural, airplane, baseball diamond, beach, buildings, chaparral, dense residential, forest, freeway, golf course, harbor, intersection, medium density residential, mobile home park, overpass, parking lot, river, runway, sparse residential, storage tanks, and tennis courts. This dataset holds highly overlapping land-use classes such as dense residential, medium residential and sparse residential, which mainly differ in the density of structures and hence make the dataset more rich and challenging. So far, this dataset is the most popular and has been widely used for the task of remote sensing image scene classification and retrieval [8, 10, 11, 14-16, 18-20, 24, 26, 28-30, 36, 38, 53, 55, 59, 61, 62, 81, 82, 84-98].

### B. WHU-RS19 Dataset

The WHU-RS19 dataset was developed in [33] on its original version of [83], extracted from a set of satellite images exported from Google Earth (Google Inc.) with spatial resolution up to 0.5 m and spectral bands of red, green, and blue. It contains 19 scene classes, including airport, beach, bridge, commercial area, desert, farmland, football field, forest, industrial area, meadow, mountain, park, parking lot, pond, port, railway station, residential area, river, and viaduct. For each scene class, there are about 50 images, with 1,005 images in total for the entire dataset. The image sizes are 600×600 pixels. This dataset is challenging due to the high variations in resolution, scale, orientation, and illuminations of the images. However, the number of images per class of this dataset is relatively small compared with UC-Merced dataset [38]. This dataset has also been widely adopted to evaluate various different scene classification methods [9, 16, 18, 33, 37, 83-87, 97].

### C. SIRI-WHU Dataset

The SIRI-WHU dataset [11] consists of 2,400 remote sensing images with 12 scene classes. Each class consists of 200 images with a spatial resolution of 2 m and a size of 200×200 pixels. It was collected from Google Earth (Google Inc.) by the Intelligent Data Extraction and Analysis of Remote Sensing (RS_IDEA) Group in Wuhan University. The 12 land-use classes contain agriculture, commercial, harbor, idle land, industrial, meadow, overpass, park, pond, residential, river, and water. Although this dataset has been tested by several methods [8, 10, 11], the number of scene classes is relatively small. Besides, it mainly covers urban areas in China and hence lacks diversity and is less challenging.

### D. RSSCN7 Dataset

The RSSCN7 dataset [17] contains a total of 2,800 remote sensing images which are composed of seven scene classes: grass land, forest, farm land, parking lot, residential region, industrial region, and river/lake. For each class, there are 400 images collected from the Google Earth (Google Inc.) that are cropped on four different scales with 100 images per scale. Each image has a size of 400×400 pixels. The main challenge of this dataset comes from the scale variations of the images [14].

### E. RSC11 Dataset

The RSC11 dataset [9] was extracted from Google Earth (Google Inc.), which covers the high-resolution remote sensing images of several US cities including Washington DC, Los Angeles, San Francisco, New York, San Diego, Chicago, and Houston. Three spectral bands were used including red, green, and blue. There are 11 complicated scene classes including dense forest, grassland, harbor, high buildings, low buildings, overpass, railway, residential area, roads, sparse forest, and storage tanks. Some of the scene classes are quite similar in vision, which increases the difficulty in discriminating the scene images. The dataset totally includes 1,232 images with about 100 images in each class. Each image has a size of 512×512 pixels and a spatial resolution of 0.2 m.

### F. Brazilian Coffee Scene Dataset

The Brazilian coffee scene dataset [82] consists of only two scene classes (coffee class and non-coffee class) and each class has 1,438 image tiles with a size of 64×64 pixels cropped from SPOT satellite images over four counties in the State of Minas Gerais, Brazil: Arceburgo, Guaranesia, Guaxupe, and Monte Santo. This dataset considered the green, red, and near-infrared bands because they are the most useful and representative ones for distinguishing vegetation areas. The identification of coffee crops was performed manually by agricultural researchers. To be specific, the creation of the dataset is performed as follows: tiles with at least 85% of coffee pixels were assigned to the coffee class; tiles with less than 10% of coffee pixels were assigned to the non-coffee class. Despite there exists significant intra-class variation caused by different crop management techniques, different plant ages, and spectral distortions, there are only two scene classes, which is quite small for testing multi-class scene classification methods.

## III. A Survey on Remote Sensing Image Scene Classification Methods

During the last decades, considerable efforts have been made to develop various methods for the task of scene classification using satellite or aerial images. As scene classification is usually carried out in feature space, effective feature representation plays an important role in constructing high-performance scene classification methods. We can generally divide existing scene classification methods into three main categories according to the features they used: handcrafted feature based methods, unsupervised feature learning based methods, and deep feature learning based methods. It should be noted that these three categories are not necessarily independent and sometimes the same method exists with different categories.

### *A. Handcrafted Feature Based Methods*

The early works for scene classification are mainly based on handcrafted features [22, 23, 27, 38, 44, 51, 56, 62, 80, 82, 99-103]. These methods mainly focus on using a considerable amount of engineering skills and domain expertise to design various human-engineering features, such as color, texture, shape, spatial and spectral information, or their combination that are the primary characteristic of a scene image and hence carry useful information used for scene classification. Here, we briefly review several most representative handcrafted features, including color histograms [99], texture descriptors [104-106], GIST [107], scale invariant feature transform (SIFT) [108], and histogram of oriented gradients (HOG) [109].

*1) Color histograms*: Among all handcrafted features, the global color histogram feature [99] is almost the simplest, yet an effective visual feature commonly used in image retrieval and scene classification [38, 56, 80, 82, 99]. A major advantage of color histograms, apart from their ease to compute, is that they are invariant to translation and rotation about the viewing axis. However, color histograms are not able to convey the spatial information, so it is very difficult to distinguish the images with the same colors but different color distribution. Besides, color histogram feature is also sensitive to small illumination changes and quantization errors.

*2) Texture descriptors*: Texture features, such as grey level co-occurrence matrix (GLCM) [104], Gabor feature [105], and local binary patterns (LBP) [84, 106, 110], etc., are widely used for analyzing aerial or satellite images [51, 56, 62, 100-102]. Texture features are commonly computed by placing primitives in local image subregions and analyzing the relative differences, so they are quite useful for identifying textural scene images.

*3) GIST*: GIST descriptor was initially proposed in [107], which provides a global description for representing the spatial structure of dominant scales and orientations of a scene. It is based on calculating the statistics of the outputs of local feature detectors in spatially distributed subregions. Specifically, in standard GIST, the images are first convoluted with a number of steerable pyramid filters. Then, the image is divided into a 4×4 grid for which orientation histograms are extracted. Note that the GIST descriptor is similar in spirit to the local SIFT descriptor [108]. Owing to its simplicity and efficiency, GIST is popularly used for scene representation [111-113].

*4) SIFT*: SIFT feature [108] describes subregions by gradient information around identified keypoints. Standard SIFT, also known as sparse SIFT, is the combination of keypoint detection and histogram based gradient representation. It generally has four steps, namely, scale space extrema searching, sub-pixel keypoint refining, dominant orientation assignment, and feature description. Except for sparse SIFT descriptor, there also exist dense SIFT that is computed in uniformly and densely sampled local regions and several extensions such as PCA-SIFT [114] and speed-up robust features (SURF) [115]. SIFT feature and its variants are highly distinctive and invariant to changes in scale, rotation, and illumination.

*5) HOG*: HOG feature was first proposed by [109] to represent objects by computing the distribution of gradient intensities and orientations in spatially distributed subregions, which has been acknowledged as one of the best features to capture the edge or local shape information of the objects. It has shown great success for many scene classification methods [22, 23, 27, 44, 103, 116, 117]. In addition, in order to further improve the description ability of HOG for remote sensing images, several extensions are also developed [118-121].

These human-engineering features have their advantages and disadvantages [56, 90, 101, 102]. In brief, the color histograms, texture descriptors, and GIST feature are global features that describe the overall statistical properties of an entire image scene in terms of certain spatial cues such as color [56, 99], texture [104-106], or spatial structure information [107], so they can be directly used by classifiers for scene classification. Whereas, SIFT descriptor and HOG feature are local features that are used for the representations of local structure [108] and shape information [109]. To represent an entire scene image, they are generally used as building blocks to construct global image features, such as the well-known bag-of-visual-words (BoVW) models [6, 8, 9, 14, 19, 29, 36, 38, 39, 55, 93, 101, 122, 123] and HOG feature-based part models [22, 23, 27, 103]. In addition, a number of improved feature encoding/pooling methods have also been proposed in the past few years, such as Fisher vector coding [10, 14, 84, 86], spatial pyramid matching (SPM) [124], and probabilistic topic model (PTM) [11, 40, 42, 43, 92, 123], etc.

In real-world applications, scene information is usually conveyed by multiple cues including spectral, color, texture, shape, and so on. Every individual cue captures only one aspect of the scene, so one single type of feature is always inadequate to represent the content of the entire scene image. Accordingly, a combination of multiple complementary features for scene classification [8, 9, 11, 12, 20, 30, 33, 85, 88, 89, 92, 125] is considered as a potential strategy to improve the performance. For example, Zhao *et al.* [11] presented a dirichlet derived multiple topic model to combine three types of features at a topic level for scene classification. Zhu *et al.* [8] proposed a local-global feature based BoVW scene classification method, in which shape-based global texture feature, local spectral feature, and local dense SIFT feature are fused.

Although the combination/fusion of multiple complementary features can often improve the results, how to effectively fuse the different types of features is still an open problem. In addition, the features introduced above are handcrafted, so the involvement of human ingenuity in feature designing significantly influences their representational capability as well as the effectiveness for scene classification. Especially when the scene images become more challenging, the description ability of those features may become limited or even impoverished.

### B. *Unsupervised Feature Learning Based Methods*

To remedy the limitations of handcrafted features, learning features automatically from images is considered as a more feasible strategy. In recent years, unsupervised feature learning from unlabeled input data has become an attractive alternative to handcrafted features and has made significant progress for remote sensing image scene classification [20, 26, 28, 33, 37, 54, 63, 87, 95, 126-134]. Unsupervised feature learning aims to learn a set of basis functions (or filters) used for feature encoding, in which the input of the functions is a set of handcrafted features or raw pixel intensity values and the output is a set of learned features. By learning features from images instead of relying on manually designed features, we can obtain more discriminative feature that is better suited to the problem at hand. Typical unsupervised feature learning methods include, but not limited to, principal component analysis (PCA) [135], *k*-means clustering, sparse coding [136], and autoencoder [137]. It is worth noting that some unsupervised feature learning models such as sparse coding and autoencoder can be easily stacked to form deeper unsupervised models.

*1) PCA*: PCA is probably the earliest unsupervised feature extraction algorithm that aims to find an optimal representative projection matrix such that the projections of the data points can best preserve the structure of the data distribution [135]. To this end, PCA learns a linear transformation matrix from input data, where the columns of the transformation matrix form a set of orthogonal basis vectors and thus make it possible to calculate the principal components of the input data. Recently, some extensions to PCA have also been proposed, such as PCANet [138] and sparse PCA [126]. In [138] PCA was employed to learn multistage filter banks, namely PCANet, which is a simple deep model without supervised training but it is able to learn robust invariant features for various image classification tasks. In [126], Chaib *et al.* presented an informative feature selection method based on sparse PCA for high resolution satellite image scene classification. However, the description power of linear PCA features is limited. It is impossible to obtain more abstract representations since the composition of linear operations yields another linear operation [139]. Some recent methods, such as sparse coding [136], autoencoder [137] that will be reviewed below, have been developed to extract nonlinear features.

*2) k-means clustering*: The *k*-means clustering is a method of vector quantization that aims to divide a collection of data items into *k* clusters, such that items within a cluster are more similar to each other than they are to items in the other clusters. Since *k*-means clustering is usually carried out when label information is not available concerning the membership of data items to predefined classes, it is seen as a typical unsupervised learning method. The algorithm iteratively repeats two steps: in the assignment step, each data item is assigned to the cluster with the closest centroid. In the update step, the centroids are recomputed as the means of the data items in the corresponding clusters. The algorithm terminates when the assignments no longer change. To initialize the algorithm, a universal method is to randomly choose *k* data items as the initial centroids. Owing to its simplicity, *k*-means clustering method is widely used for unsupervised feature learning-based scene image classification. The most representative example is BoVW-based methods [8, 9, 14, 19, 29, 86, 87, 89, 91, 122, 123, 132, 140] where the visual dictionaries (codebooks) are generated by performing *k*-means clustering on the set of local features.

*3) Sparse coding*: Recently, sparse coding [136] instead of vector quantization has been applied to dictionary learning. It is a class of unsupervised method for learning an over-complete dictionary from unlabeled training samples, such that we can represent an input sample as a linear combination of these atoms efficiently [136]. The coefficient vectors [26, 33, 37, 95, 128] or reconstruction residuals [20, 47] for each sample form their new feature representations. The core of sparse coding is to encode each high-dimensional input vector sparsely by a few structural primitives in a low-dimensional manifold [141]. The procedure of seeking the sparsest representation for a sample in terms of the over-complete dictionary endows itself with a discriminative nature to perform classification. Learning a set of basis vectors using sparse coding consists of two separate optimizations: the first is an optimization over sparse coefficients for each training sample and the second is an optimization over basis vectors across a batch of training samples at once.

More recently, a large number of sparse coding methods have been proposed for scene classification of remote sensing images [20, 26, 28, 33, 37, 95, 128]. For example, [26] adopted sparse coding to learn a set of basis functions from unlabeled low-level features. The low-level features were then encoded in terms of the basis functions to generate spare feature representations. [33] introduced a sparse coding based multiple feature combination for satellite scene classification. [28] designed a novel method for satellite image annotation using multi-feature joint sparse coding with spatial relation constraint. However, sparse coding is computationally expensive when dealing with large-scale data. Inspired by the observation that nonzero coefficients are usually assigned to nearby elements in the dictionary, Wang *et al.* [142] proposed locality constrained linear coding (LLC) method. LLC assumes that a data can be reconstructed by using its *k*-nearest neighbors in the dictionary. Thus, the sparse coefficients can be computed through solving a least-squares problem. The weights for the remaining atoms are set to zero and the sparsity is replaced with locality [85, 142].

*4) Autoencoder*: Autoencoder [137] is another famous

unsupervised feature learning method that has been successfully applied to land use scene classification [63, 129, 133, 134]. It is a symmetrical neural network that is used to learn a compressed feature representation from high dimensional feature space in an unsupervised manner. This is achieved by minimizing the reconstruction error between the input data at the encoding layer and its reconstruction at the decoding layer. The number of nodes in the encoding layer is equal to that of the decoding layer. To reduce the dimensionality of data, the autoencoder network reconstructs the feature representation with fewer nodes in the hidden layers. The activations of the hidden layer are usually used as the compressed features. Gradient decent with back propagation is used for training the networks. To train a multilayer stacked autoencoder, the most important issue is how to initialize the networks. If initial weights are large, autoencoder tends to converge to poor local minima, while small initial weights lead to tiny gradient in early layer, making it infeasible to train such a multilayer network. Fortunately, a pre-training approach was proposed by Hinton *et. al*. [137], which found that initializing weights using restricted Boltzmann machines (RBM) is closer to a good solution.

In real applications, the aforementioned unsupervised feature learning methods have achieved good performance for land use classification, especially compared to handcrafted feature based methods. However, the lack of semantic information provided by the category labels can not guarantee the best discrimination ability between classes because unsupervised feature learning methods do not make use of data class information. For a better classification performance, we still need to use labeled data to develop supervised feature learning methods, which will be reviewed below, to extract more powerful features.

### *C. Deep Feature Learning Based Methods*

Most of the current state-of-the-art approaches generally rely on supervised learning to obtain good feature representations. Especially in 2006, a breakthrough in deep feature learning was made by Hinton and Salakhutdinov [137]. Since then, the aim of researchers has been to replace hand-engineered features with trainable multilayer networks and a amount of deep learning models have shown impressive feature representation capability for a wide range of applications including remote sensing image scene classification [13, 17, 45, 50, 82, 143-158].

On the one hand, in comparison with traditional handcrafted features that require a considerable amount of engineering skill and domain expertise, deep learning features are automatically learned from data using a general-purpose learning procedure via deep-architecture neural networks. This is the key advantage of deep learning methods. On the other hand, compared with aforementioned unsupervised feature learning methods that are generally shallow-structured models (e.g., sparse coding), deep learning models that are composed of multiple processing layers can learn more powerful feature representations of data with multiple levels of abstraction [159]. In addition, deep feature learning methods have also turned out to be very good at discovering intricate structures and discriminative information hidden in high-dimensional data, and the features from toper layers of the deep neural network show semantic abstracting properties. All of these make deep features more applicable for semantic-level scene classification.

Currently, there exist a number of deep learning models, such as deep belief nets (DBN) [160], deep Boltzmann machines (DBM) [161], stacked autoencoder (SAE) [162], convolutional neural networks (CNNs) [163-167], and so on. Limited by the space, here we mainly review two widely used deep learning methods including SAE and CNN. Readers can refer to relevant publications for other deep learning methods.

*1) SAE*: SAE [162] is a relatively simple deep learning model that has been successfully applied for remote sensing image scene classification [13, 134]. An SAE consists of multiple layers of autoencoders in which the outputs of each layer are wired to the inputs of the successive layer. To train an SAE, a feasible way is to use greedy layer-wise training scheme [168]. Specifically, one should first train the first layer on raw input data to obtain parameters and transfer the raw data into an intermediate vector consisting of activations of the hidden units. Then, this process is repeated for subsequent layers by using the output of each layer as input for the subsequent layer. This method trains the parameters of each layer individually while freezing parameters for the remainder of the model. To obtain better results, after layer-wise training is completed, fine-tuning is performed to tune the parameters of all layers at the same time with a smaller learning rate. Compared to a single autoencoder as mentioned in previous subsection, the feature representation power of SAE can be significantly strengthened. This can be easily explained: with the composition of multiple autoencoder that each transforms the representation at one level (starting with the raw input) into a representation at a higher, slightly more abstract level, we can learn very powerful representations. This has been proven in literatures [13, 134, 169-171].

*2) CNNs:* CNNs are designed to process data that come in the form of multiple arrays, for example a multi-spectral image composed of multiple 2D arrays containing pixel intensities in the multiple band channels. Starting with the impressive success of AlexNet [163], many representative CNN models including Overfeat [164], VGGNet [165], GoogLeNet [166], SPPNet [167], and ResNet [172] have been proposed in the literature. There exist four key ideas behind CNNs that take advantage of the properties of natural signals, namely, local connections, shared weights, pooling, and the use of many layers [159].

The architecture of a typical CNN is structured as a series of layers. (i) *Convolutional layers*: They are the most important ones for extracting features from images. The first layers usually capture low-level features (like edges, lines and corners) while the deeper layers are able to learn more expressive features (like structures, objects and shapes) by combining low-level ones. (ii) *Pooling layers*: Typically, after each convolutional layer, there exist pooling layers that are created by computing some local non-linear operation of a particular feature over a region of the image. This process ensures that the same result can be obtained, even when image features have small translations or rotations, which is very important for scene classification and detection.

Table 1. Comparison between the proposed NWPU-RESISC45 dataset and some other publicly available datasets. Our dataset provides many more images per class, scene classes, total images, and scales (spatial resolution variations) in comparison with other available datasets for remote sensing image scene classification.

| Datasets | Images per class | Scene classes | Total images | Spatial resolution (m) | Image sizes | Year |
|---|---|---|---|---|---|---|
| UC Merced Land-Use [38] | 100 | 21 | 2,100 | 0.3 | 256×256 | 2010 |
| WHU-RS19 [33] | ~50 | 19 | 1,005 | up to 0.5 | 600×600 | 2012 |
| SIRI-WHU [11] | 200 | 12 | 2,400 | 2 | 200×200 | 2016 |
| RSSCN7 [17] | 400 | 7 | 2,800 | -- | 400×400 | 2015 |
| RSC11 [9] | ~100 | 11 | 1,232 | 0.2 | 512×512 | 2016 |
| Brazilian Coffee Scene [82] | 1438 | 2 | 2,876 | -- | 64×64 | 2015 |
| **NWPU-RESISC45** | **700** | **45** | **31,500** | **~30 to 0.2** | **256×256** | **2016** |

(iii) *Normalization layers*: They aim to improve generalization inspired by inhibition schemes presented in the real neurons of the brain. (iv) *Fully connected layers*: They are typically used as the last few layers of the network. By removing constraints, they can better summarize the information conveyed by lower-level layers in view of the final decision. As a fully-connected layer occupies most of the parameters, over-fitting can easily happen. To prevent this, the dropout method was employed [163].

In recent years, there has been an extensive popularity of CNNs in various remote sensing applications, such as geospatial object detection [45, 50, 173] and land use scene classification [82, 143, 144, 146-150, 152, 153]. For instance, to address the problem of object rotation variations, [45] proposed a novel approach to learn a rotation-invariant CNN (RICNN) model for boosting the performance of object detection on the basis of the existing CNN models. This is achieved by optimizing a new objective function via imposing a regularization constraint term, which explicitly enforces the feature representations of the training samples before and after rotation to be mapped close to each other. [152] and [153] explored the use of existing CNNs for scene classification of remote sensing images by using three different strategies: full training, fine tuning, and using CNNs as feature extractors. Experimental results show that fine tuning tends to be the best performing strategy on small-scale datasets.

## IV. The Proposed NWPU-RESISC45 Dataset

Since the emergence of UC Merced land use dataset [38], considerable efforts have been dedicated toward the construction of various datasets [9, 11, 17, 33, 38, 82] for the task of remote sensing image scene classification. Despite the remarkable progress made so far, as we reviewed in Section II and summarized in Table 1, almost all existing remote sensing image datasets have a number of notable limitations, such as the small scale of scene classes, the image numbers per class and the total image numbers, the lack of scene variations and diversity, and the saturation of classification accuracy (e.g., ~100% accuracy on the most popular UC Merced dataset with deep CNN features [82]). These limitations have severely limited the development of new data driven algorithms and also prohibited the wide use of deep learning methods because almost all deep learning models are required to be trained on large training datasets with abundant and diverse images to avoid over-fitting.

Under such a circumstance, proposing a large-scale dataset with big image variations and diversity is highly desirable due to the potential benefits for remote sensing community. This motivates us to propose the NWPU-RESISC45 dataset, which is a freely and publicly available benchmark dataset used for remote sensing image scene classification.

### A. Selecting Scene Classes for NWPU-RESISC45 Dataset

The first step of constructing the dataset is to select a list of representative scene classes. To this end, we first investigated all scene classes of the existing datasets [9, 11, 17, 33, 38, 82] to form 30 widely-used scene categories. Then, we searched the keywords of "OBIA", "GEOBIA", "land cover classification", "land use classification", "geographic image retrieval", and "geospatial object detection" on Web of Science and Google Scholar to carefully select 15 meaningful scene classes. In such a case, we obtained a total of 45 scene classes to construct the proposed NWPU-RESISC45 dataset. These 45 scene classes are as follows: airplane, airport, baseball diamond, basketball court, beach, bridge, chaparral, church, circular farmland, cloud, commercial area, dense residential, desert, forest, freeway, golf course, ground track field, harbor, industrial area, intersection, island, lake, meadow, medium residential, mobile home park, mountain, overpass, palace, parking lot, railway, railway station, rectangular farmland, river, roundabout, runway, sea ice, ship, snowberg, sparse residential, stadium, storage tank, tennis court, terrace, thermal power station, and wetland.

Here, it should be pointed out that the term of "scene" used for this scene classification dataset refers to its generalized form, including land use and land cover classes (e.g., commercial area, farmland, forest, industrial area, mountain, residential area), man-made object classes (e.g., airplane, airport, bridge, church, palace, ship), as well as landscape nature object classes (e.g., beach, cloud, island, lake, river, sea ice). Accordingly, these classes contain a variety of spatial patterns, some homogeneous with respect to texture, some homogeneous with respect to color, others not homogeneous at all.

### B. The NWPU-RESISC45 Dataset

The proposed NWPU-RESISC45 dataset consists of 31,500 remote sensing images divided into 45 scene classes. Each class includes 700 images with a size of 256×256 pixels in the red green blue (RGB) color space. The spatial resolution varies from about 30 m to 0.2 m per pixel for most of the scene classes except for the classes of island, lake, mountain, and snowberg that have lower spatial resolutions. As same as most of the existing datasets [9, 11, 17, 33], this dataset was also extracted, by the experts in the field of remote sensing image interpretation, from Google Earth (Google Inc.) that maps the Earth by the superimposition of images obtained from satellite imagery, aerial photography and geographic information system (GIS) onto a 3D globe. The 31,500 images cover more than 100 countries and regions all over the world, including developing, transition, and highly developed economies. Figure 2 shows two samples of each class from this dataset.

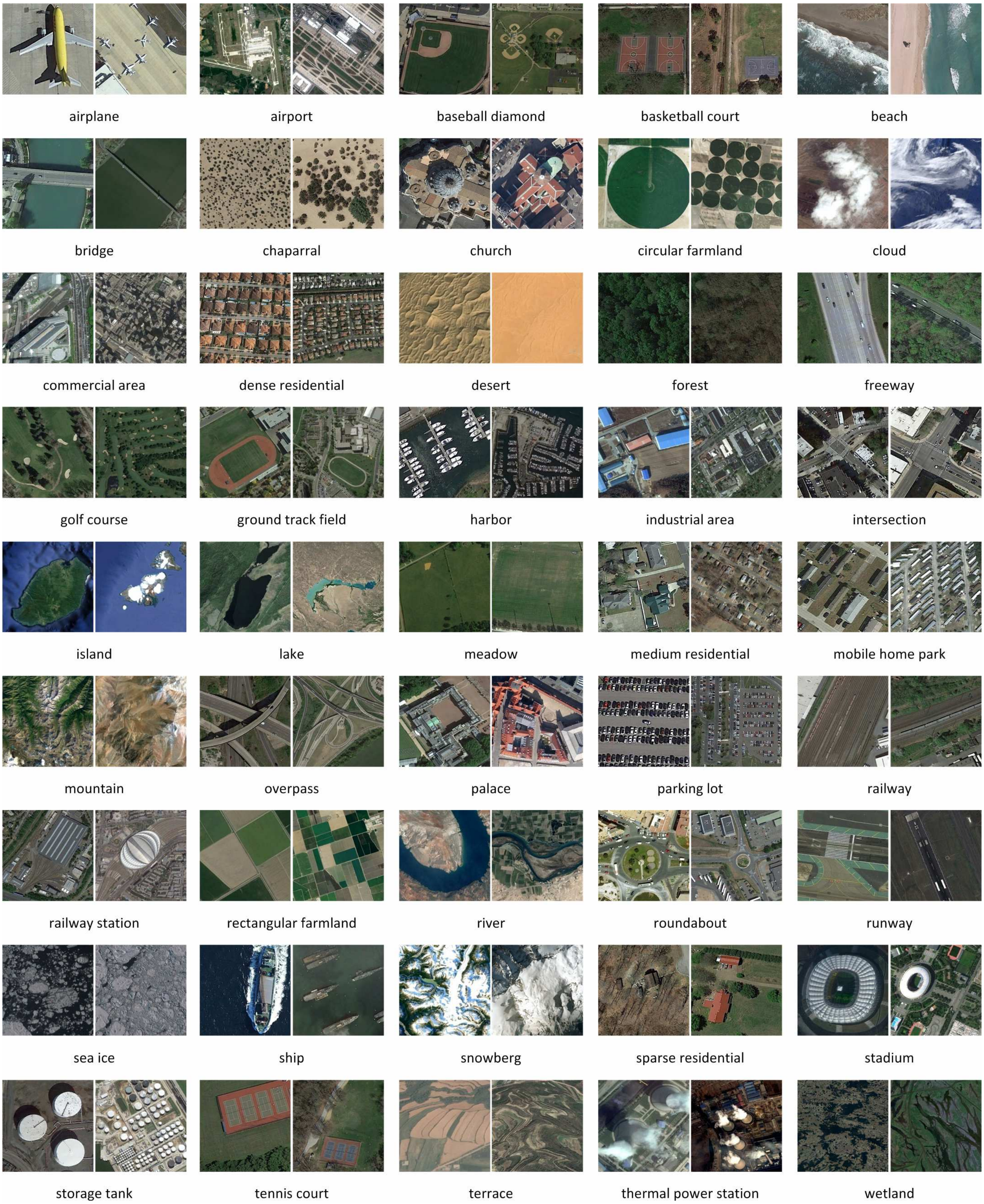

Figure 2. Some example images from the proposed NWPU-RESISC45 dataset, which was carefully designed under all kinds of weathers, seasons, illumination conditions, imaging conditions, and scales. Accordingly, these images generally have rich variations in translation, viewpoint, object pose and appearance, spatial resolution, illumination, background, and occlusion, etc.

The proposed NWPU-RESISC45 dataset has the following three notable characteristics compared with all existing scene classification datasets including [9, 11, 17, 33, 38, 82].

*1) Large scale:* The resurrection of deep learning has had a revolutionary impact on the current state-of-the-arts in machine learning and computer vision. A major contributing factor to its success is the availability of large-scale datasets that allows deep networks to develop their full potential. Unfortunately, as can be seen from Table 1, almost all existing datasets are small scale. To address this problem, we propose a large-scale, freely and publicly available benchmark dataset, which covers 31,500 images and 45 scene classes with 700 images per class. Taking the most widely-used UC Merced dataset [38] as comparison, the total image number of our dataset is 15 times larger than it. To the best of our knowledge, our dataset is of the largest scale on the number of scene classes and the total image number. The creation of this new dataset will enable the community to develop and test various data-driven algorithms to further boost the state-of-the-arts.

*2) Rich image variations:* Tolerance to image variations is an important desired property of any scene classification system, be it human or machine. However, most of the existing datasets are not very rich in terms of image variations. On the contrary, our images were carefully selected under all kinds of weathers, seasons, illumination conditions, imaging conditions, and scales. Thus, for each scene category, our dataset possesses much rich variations in translation, viewpoint, object pose and appearance, spatial resolution, illumination, background, and occlusion, etc.

*3) High within class diversity and between class similarity:* Many top-performing methods built upon deep neural networks have achieved saturation of classification accuracy on most of the existing datasets owing to their simplicity, or rather the lack of variations and diversity. With this in mind, our new dataset is rather challenging with high within class diversity and between class similarity. To this end, we obtained images under all kinds of conditions and added some more fine-grained scene classes with high semantic overlapping such as circular farmland and rectangular farmland, commercial area and industrial area, basketball court and tennis court, and so on.

# V. BENCHMARKING REPRESENTATIVE METHODS

Current methods have only been evaluated on small datasets. It is unclear how they perform on a large-scale dataset. In this section we evaluate a number of representative approaches for scene classification on our NWPU-RESISC45 dataset.

## A. *Representative Methods*

A total of 12 kinds of previous and current state-of-the-art features, which have been widely used for scene classification, are selected for evaluation. Specifically, our selections include three global handcrafted features: color histograms, LBP, GIST, three unsupervised feature learning models: dense SIFT based BoVW feature and its two extensions (BoVW+SPM and LLC), three deep learning-based CNN features: AlexNet, VGGNet-16, GoogLeNet, and three fine-tuned CNN features.

*1) Color histograms*: Color histograms feature [99] is almost the simplest handcrafted feature that has been widely used for image classification. In our work, the color histograms feature is directly computed in RGB color space because of its simplicity. Each channel is quantized into 64 bins for a total histogram feature length of 192. The histograms are normalized to have an L1 norm of one.

*2) LBP*: LBP feature [106] is a theoretical simple yet efficient approach for texture description by computing the frequencies of local patterns in subregions. For an image, it first compares each central pixel to its $N$ neighbors: when the neighbor's value is bigger than the value of center pixel, output 1, otherwise, output 0. This forms an $N$-bit decimal number to describe each center pixel. The LBP descriptor is obtained by computing the histogram of the decimal numbers over the image and results in a feature vector with $2^N$ dimensions. In our implementation, we set $N$=8, and hence resulting in a 256 dimensional LBP feature.

*3) GIST*: GIST feature [107] describes the global spatial layout of a scene image. To compute GIST feature, each image is first decomposed by a bank of multiscale-oriented Gabor filters (set to 8 orientations and 4 different scales as the original work of [107]). The result of each filter is then averaged over 16 non-overlapping regions arranged on a 4×4 grid. The resulting image representation is a 512 dimensional feature vector.

*4) BoVW*: BoVW [174] is possibly one of the most popular visual features during the last decade. Owing to its efficiency and invariance to viewpoint changes, it has been widely used by the community for geographic image classification. There are usually five major steps in the BoVW framework used for image classification. (i) *Patch extraction*. With an image as input, the outputs of this step are image patches. This step is implemented via sampling local areas of images in a dense or sparse manner. (ii) *Patch representation*. Given image patches, the outputs of this step are their feature descriptors such as the popular SIFT descriptor. (iii) *Codebook generation*. The inputs of this step are feature descriptors extracted from all training images and the output is a visual codebook. The codebook is usually formed by unsupervised $k$-means clustering over all feature descriptors, as described in subsection III.B. (iv) *Feature encoding*. Given feature descriptors and codebook as input, this step quantizes each feature descriptor into a visual word in the codebook. (v) *Feature pooling*. This step pools encoded local descriptors into a global histogram representation for each image.

Table 2. Parameters utilized for CNN model fine-tuning.

| CNN | #iterations | Batch size | Leaning rate | Leaning rate (last layer) | Weight decay | Momentum |
|---|---|---|---|---|---|---|
| AlexNet [163] | 15000 | 128 | 0.001 | 0.01 | 0.0005 | 0.9 |
| VGGNet-16 [165] | 15000 | 50 | 0.001 | 0.01 | 0.0005 | 0.9 |
| GoogLeNet [166] | 15000 | 128 | 0.001 | 0.01 | 0.0005 | 0.9 |

*5) BoVW+SPM*: SPM [124] is used to incorporate spatial information of a scene image. In brief, SPM divides each image into increasingly finer spatial subregions, constructs the BoVW representations for each subregion, and then concatenates them to represent the image. In our implementation, we divide each image into 1×1 and 2×2 subregions. Thus, given a codebook with the size $K$, we can obtain a $5K$ dimensional feature vector for each image by using BoVW+SPM scheme.

*6) LLC*: LLC [142] is an integrated variation of sparse coding and BoVW. It utilizes the locality constraints to project each feature descriptor into its local-coordinate neighbors, and the projected coordinates are integrated by max pooling to form the final representation. Note that in our implementation of LLC, we adopted the same codebook as BoVW that was constructed simply by $k$-means clustering without optimization. Therefore, LLC and BoVW have the same feature dimension.

*7) AlexNet*: AlexNet [163] was first proposed by Alex Krizhevsky *et al*. and was the winner of ImageNet large scale visual recognition challenge (ILSVRC) [175] in 2012. It is composed of five convolutional layers and three fully connected layers. Besides, response normalization layers follow the first and the second convolutional layers. Max-pooling layers follow both response normalization layers and the fifth convolutional layer. It is a landmark study for machine learning and computer vision since it was the first work to employ non-saturating neurons, GPU implementation of the convolution operation and dropout to prevent over-fitting. In our evaluation, we extracted the AlexNet CNN feature from the second fully connected layer, which results in a feature vector of 4,096 dimensions.

*8) VGGNet-16*: VGGNet was proposed in [165] and has won the localization and classification tracks of the ILSVRC-2014 competition. It has two famous architectures: VGGNet-16 and VGGNet-19. In this evaluation, we used the former one because of its simpler architecture and slightly better performance. It has 13 convolutional layers, 5 pooling layers, and 3 fully connected layers. The VGGNet-16 CNN feature was also extracted from the second fully connected layer to obtain a feature vector of 4,096 dimensions.

*9) GoogLeNet*: GoogLeNet [166] is another representative CNN architecture that achieved new state of the art for the task of classification and detection in the ILSVRC-2014. The main hallmark of its architecture is the improved utilization of the computing resources inside the network. By a carefully crafted design, the depth and width of the network were increased while keeping the computational budget constant. GoogLeNet has two main advantages: (i) utilization of filters of different sizes at the same layer, which maintains more spatial information, and (ii) reduction of the number of parameters of the network, making it less sensitive to over-fitting and allowing it to be deeper. The 22-layer GoogLeNet has more than 50 convolutional layers distributed inside the inception modules. However, in fact, GoogLeNet has 12 times fewer parameters than AlexNet. In our work, we extracted the GoogLeNet CNN feature from the last pooling layer to form a feature vector of 1,024 dimensions.

*10)-12) Fine-tuned AlexNet, VGGNet-16, and GoogLeNet*: Except for using the aforementioned three CNNs as universal feature extractors, we also fine-tuned them on our new dataset to obtain better performance without using any data augmentation technique. Table 2 summarizes the detailed parameters used for fine-tuning. Here, we used a bigger learning rate (0.01) for the last layer to make the model be able to jump out of local optimum and to converge fast and a smaller learning rate (0.001) for other layers to allow fine-tuning to make progress while not clobbering the initialization.

### *B. Experimental Setup*

To make a comprehensive evaluation, two training-test ratios are considered. (i) 10%-90%: the dataset was randomly split into 10% for training and 90% for testing (70 training samples and 630 testing samples per class). (ii) 20%-80%: the dataset was randomly divided into 20% for training and 80% for testing (140 training samples and 560 testing samples per class).

For BoVW, BoVW+SPM and LLC methods, we adopted the widely used densely sampled SIFT descriptor to describe each image patch with the patch size set to be 16×16 pixels and the grid spacing to be 8 pixels to balance the speed/accuracy trade-off, which is the same as the research work of [86]. The sizes of visual codebooks were set to be 500, 1000, 2000, and 5000, respectively, to study how they affected the classification performance.

The AlexNet model, VGGNet-16 model, and GoogLeNet model, which were pre-trained on ImageNet dataset [175], are obtained from https://github.com/BVLC/caffe/wiki/Model-Zoo for deep CNN feature extraction. To further improve their generalization capability, we also fine-tuned them by using the parameters as summarized in Table 2. All three CNN models were implemented on a PC with 2 2.8GHz 6-core CPUs and 32GB memory. Besides, a GTX Titan X GPU was also used for acceleration.

For fair comparison, the image classification was carried out by using linear support vector machines (SVMs) for all 12 kinds of image features. We implemented it with the LibSVM toolbox [176] and used the default setting in linear SVM ($C$=1) to tune the trade-off between the amount of accepted errors and the maximization of the margin. Specifically, after extracting the aforementioned 12 kinds of image features, a linear one-*v.s.*-all SVM classifier was trained for each scene class by treating the images of the chosen class as positives and the rest images as negatives. An unlabeled test image is assigned to label of the classifier with the highest response.

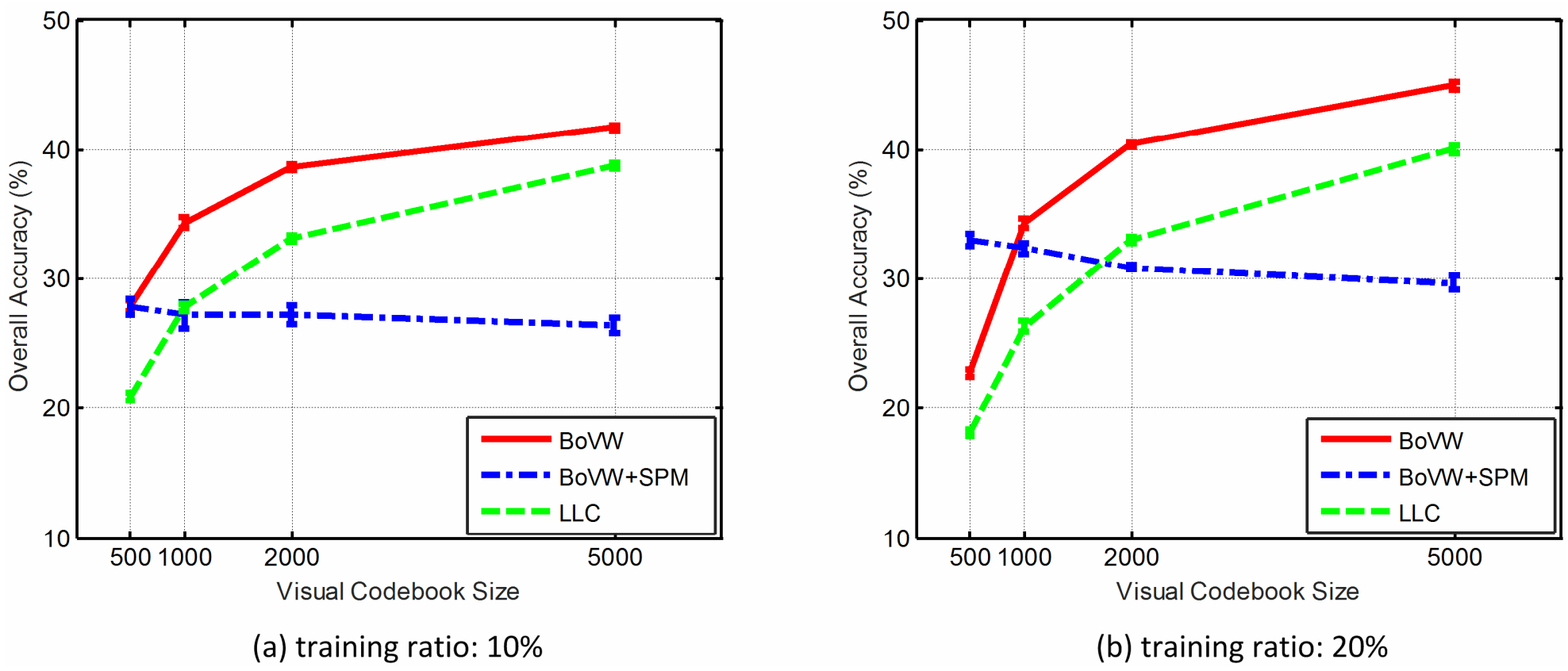

Figure 3. Overall accuracies of the methods of BoVW, BoVW+SPM and LLC with the visual codebook sizes being set to be 500, 1000, 2000, and 5000, respectively, under the training ratios of (a) 10% and (b) 20%.

Table 3. Overall accuracies (%) of three kinds of handcrafted features under the training ratios of 10% and 20%.

| Features | Training ratios | |
|---|---|---|
| | 10% | 20% |
| Color histograms | 24.84±0.22 | 27.52±0.14 |
| LBP | 19.20±0.41 | 21.74±0.18 |
| GIST | 15.90±0.23 | 17.88±0.22 |

Table 4. Overall accuracies (%) of three kinds of unsupervised feature learning methods under the training ratios of 10% and 20%.

| Features | Training ratios | |
|---|---|---|
| | 10% | 20% |
| BoVW | 41.72±0.21 | 44.97±0.28 |
| BoVW+SPM | 27.83±0.61 | 32.96±0.47 |
| LLC | 38.81±0.23 | 40.03±0.34 |

Table 5. Overall accuracies (%) of three kinds of deep learning-based CNN features under the training ratios of 10% and 20%.

| Features | Training ratios | |
|---|---|---|
| | 10% | 20% |
| AlexNet | 76.69±0.21 | 79.85±0.13 |
| VGGNet-16 | 76.47±0.18 | 79.79±0.15 |
| GoogLeNet | 76.19±0.38 | 78.48±0.26 |

Table 6. Overall accuracies (%) of three kinds of fine-tuned deep CNN features under the training ratios of 10% and 20%.

| Features | Training ratios | |
|---|---|---|
| | 10% | 20% |
| Fine-tuned AlexNet | 81.22±0.19 | 85.16±0.18 |
| Fine-tuned VGGNet-16 | 87.15±0.45 | 90.36±0.18 |
| Fine-tuned GoogLeNet | 82.57±0.12 | 86.02±0.18 |

### *C. Evaluation Metrics*

There exist three widely-used, standard evaluation metrics in image classification: overall accuracy, average accuracy, and confusion matrix. The overall accuracy is defined as the number of correctly classified samples, regardless of which class they belong to, divided by the total number of samples. The average accuracy is defined as the average of the classification accuracy of each class, regardless of the number of samples in each class. The confusion matrix is an informative table used to analyze all the errors and confusions between different classes which is generated by counting each type of correct and incorrect classification of the test samples and accumulating the results in the table.

Here, it should be pointed out that our dataset has the same image number per class, so the value of overall accuracy equals to the value average accuracy. Thus, in this paper, we just used the metrics of overall accuracy and confusion matrix to evaluate all classification methods. In addition, in order to obtain reliable results for the metrics of overall accuracy and confusion matrix, we repeated the experiment ten times for each training-test ratio and report the mean and standard deviation of the results.

### *D. Experimental Results*

Figure 3 first presents the overall accuracies of the methods of BoVW, BoVW+SPM and LLC with the visual codebook sizes being set to be 500, 1000, 2000, and 5000, respectively, under the training ratios of 10% and 20%. As can be seen from it, the size of codebook affects the accuracy remarkably and (i) for the training ratio of 10%, the best results are obtained with the codebook sizes of 5000, 500, 5000, and (ii) for the training ratio of 20%, the best results are obtained with the codebook sizes of 5000, 500, 5000, for BoVW, BoVW+SPM and LLC methods, respectively. Consequently, in our subsequent evaluations, the results of BoVW, BoVW+SPM and LLC are all based on these optimal parameter settings.

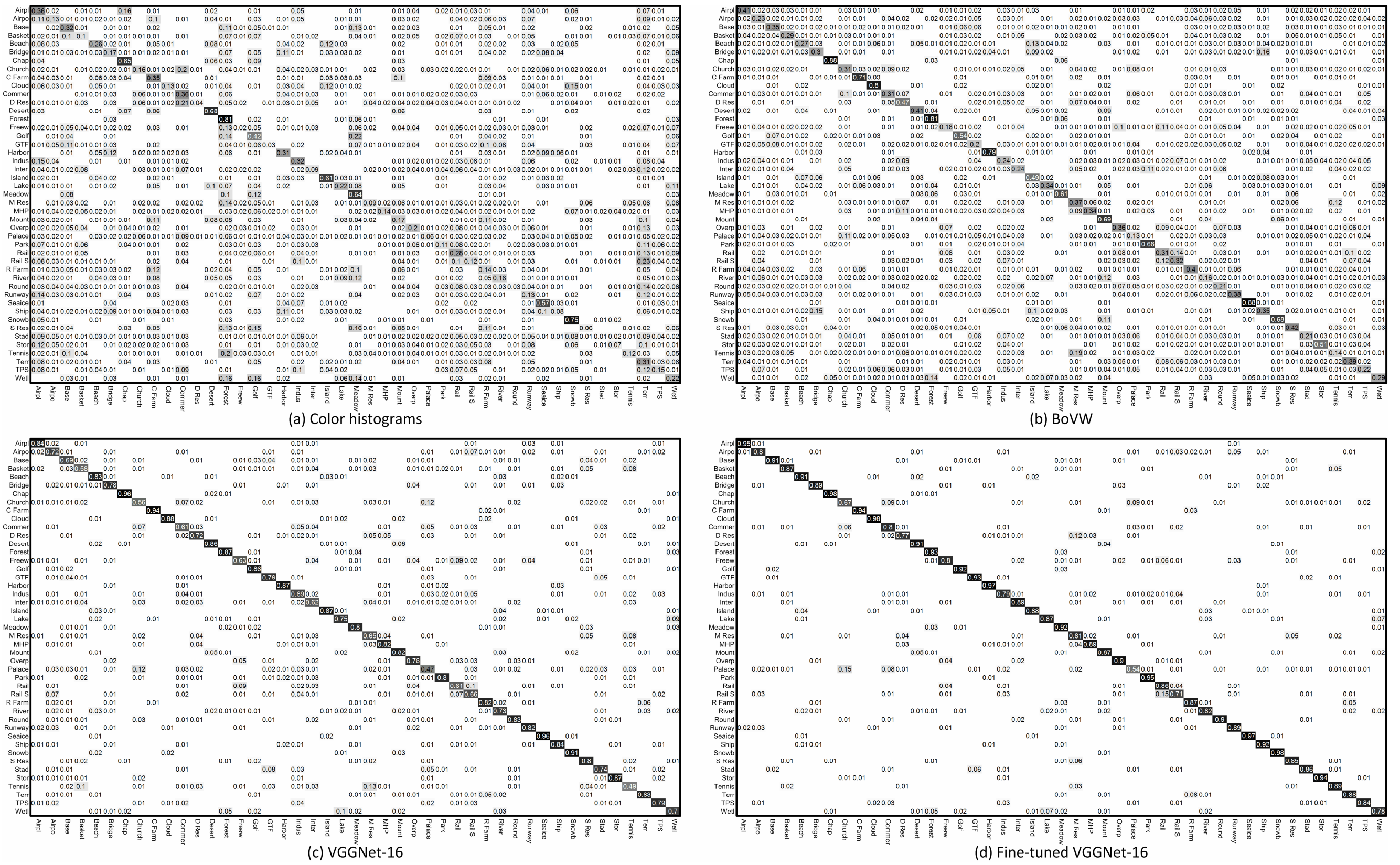

Figure 4. Confusion matrices under the training ratio of 10% by using the following methods: (a) Color histograms, (b) BoVW, (c) VGGNet-16, and (d) Fine-tuned VGGNet-16.

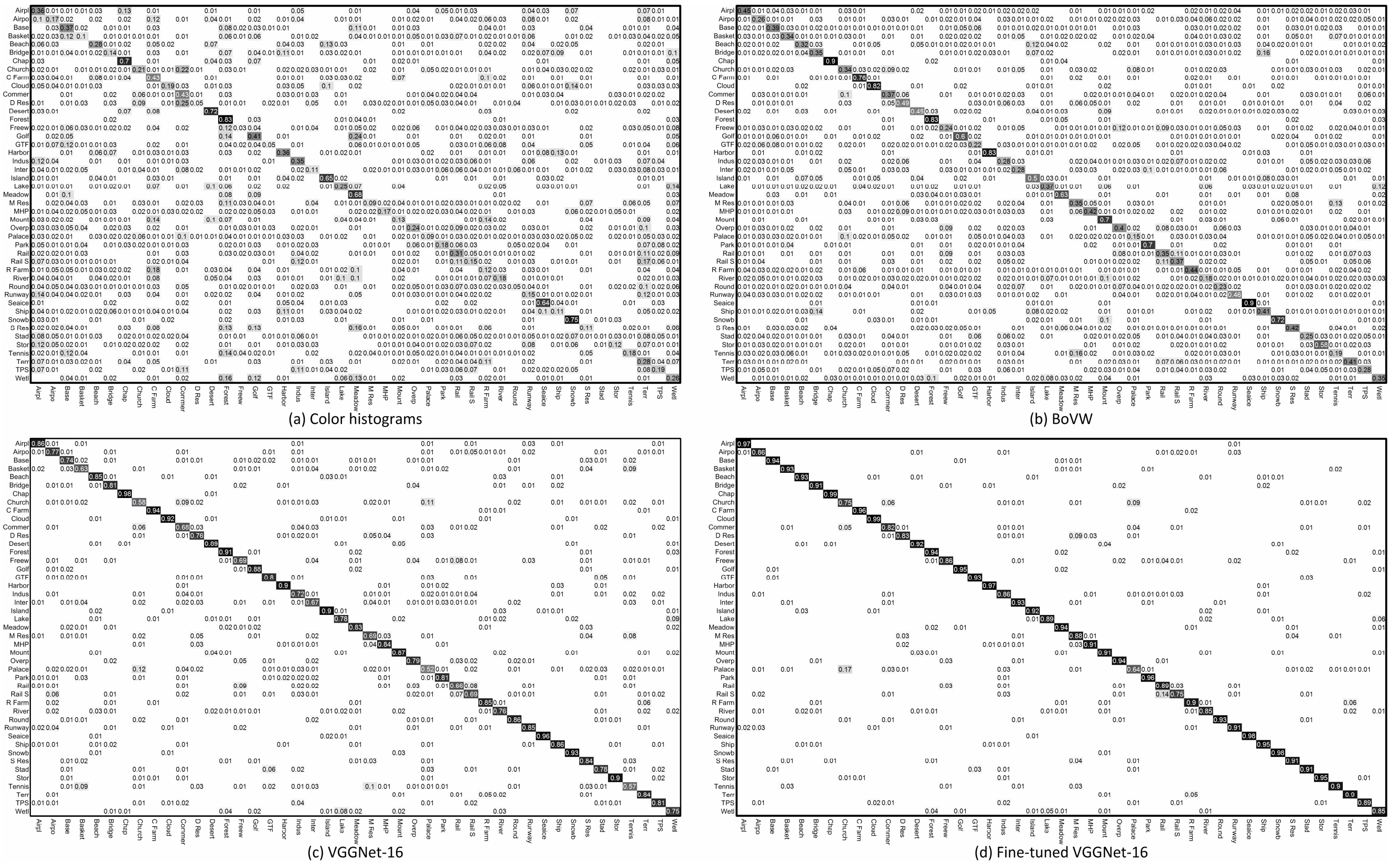

Figure 5. Confusion matrices under the training ratio of 20% by using the following methods: (a) Color histograms, (b) BoVW, (c) VGGNet-16, and (d) Fine-tuned VGGNet-16.

Tables 3-6 show the overall accuracies of three handcrafted global features, three unsupervised feature learning methods, three deep CNN features, and three fine-tuned CNN features, respectively, under the training ratios of 10% and 20%. As can be seen from Tables 3-6: (i) Handcrafted low-level features have the relatively lowest classification accuracy. (ii) BoVW and its two variants, namely BoVW+SPM and LLC, improve significantly handcrafted low-level features. Actually, they act as mid-level image features that are built on low-level features such as SIFT descriptor [108] and hence provide more semantic and more robust representations than the low level ones for filling-up the so-called semantic-gap. (iii) Deep CNN features outperform all handcrafted features and unsupervised feature learning methods in very big margins (at least 30% performance improvement). This demonstrates the huge superiority of the current-dominated deep learning methods in comparison with previous state-of-the-art methods. (iv) By fine-tuning the three off-the-shelf CNN models, the accuracy was further boosted by at least six percentage points, resulting in the highest accuracy. Figures 4-5 show the confusion matrices of different methods under the training ratios of 10% and 20%, respectively, where the entry in the $i$-th row and $j$-th column denotes the rate of test samples from the $i$-th class that are classified as the $j$-th class. Limited by the space, we here just report the confusion matrices with the highest overall accuracies selected from each class of features (handcrafted features, unsupervised feature learning methods, CNN features, and fine-tuned CNN features). From Figures 4-5, we observed that: (i) In agreement with the overall accuracies as shown in Tables 3-6, handcrafted features have the lowest per-class accuracies, unsupervised feature learning methods take the second place, and deep learning based CNN features have the highest per-class accuracies. (ii) For color histogram feature, the relatively big confusions happen between 'golf court' and 'meadow' because they are characterized by green color. (iii) For BoVW and deep learning based CNN features, the relatively big confusions happen between 'church' and 'palace' and 'dense residential' and 'medium residential' because of their similar global structure or spatial layout. This suggests that a potential way to classify more challenging image scenes may be deep learning based methods in combination with discriminative attributes oriented methods such as [23].

## VI. Conclusion and Discussion

The significant development of remote sensing technology over the past decade has been providing us explosive remote sensing data for intelligent earth observation such as scene classification using remote sensing images. However, the lack of publicly available "big data" of remote sensing images severely limits the development of new approaches especially deep learning based methods. This paper first presented a comprehensive review of the recent progress in the field of remote sensing image scene classification, including benchmark datasets and state-of-the-art methods. Then, by analyzing the limitations of the existing datasets, we proposed a large-scale, freely and publicly available benchmark dataset to enable the community to develop and evaluate new data-driven algorithms. Finally, we evaluated a number of representative state-of-the-art methods including deep learning based methods for the task of scene classification using the proposed dataset and reported the results as a useful performance baseline for future research.

However, until now, almost all scene classification methods rely on only traditional remote sensing, i.e., overhead imagery, to distinguish different types of land-cover in a given region. In fact, the more recent development of social media and spatial technology has significant potential for complimenting the shortcoming of traditional means of scene classification. First, the rapid development of social media especially the on-line photo sharing websites, such as Flickr, Panoramio, and Instagram, have been collecting sorts of information of ground objects from geo-tagged ground photo collections. Second, by linking earth observation data coming from satellites and geographic information systems (GIS) to location-aware spatial technologies, such as global positioning system (GPS), wireless fidelity (WIFI), and smart-phones, we are locating at a powerful geographic information system in which we can readily know, at any time, where every ground object is located on the surface of the Earth. Therefore, we can say that these geo-tagged ground photo collections and location-based geographic resources act as a repository of all kinds of information, including who, what, where, when, why, and how. This allows us to perform knowledge discovery by crowdsourcing of information through these location-based social medial data. For example, using only remote sensing images, it is more difficult to tell where a certain object comes from, but with Twitter generating more than 500 million Tweets per day (of which a good portion are tagged with latitude-longitude coordinates), we can map "what-is-where" easily on the surface of the Earth using the "what" and "where" aspects of the information. Besides, compared with remote sensing images, the ground photos uploaded by user holds higher resolution and are quite different from satellite remote sensing in the observation direction, which can well capture the detail and vertical characteristics of ground objects. All the additional information is in fact very useful for the classification and recognition of remote sensing images because it could better help individuals learn more powerful (or multi-view) feature representations. Consequently, in the future work we need to explore new methods and systems in which the combination of remote sensing data and information coming from social media and spatial technology can be deployed to promote the state-of-the-art of remote sensing image scene classification.